\documentclass[conference]{IEEEtran}
\IEEEoverridecommandlockouts

\usepackage{amsmath,amssymb,bm}
\usepackage{booktabs}
\usepackage{graphicx}
\usepackage{xcolor}
\usepackage{algorithm}
\usepackage{algpseudocode}
\usepackage{microtype}
\usepackage{url}
\usepackage{eso-pic}
\usepackage{tikz}
\usetikzlibrary{arrows.meta,calc}
\usepackage{adjustbox}
\usetikzlibrary{arrows.meta,positioning,calc,fit,backgrounds}
\newcommand{\ind}{\mathbf{1}}
\newcommand{\agents}{\mathcal A_t}

\newcommand{\grid}{\Lambda_N}
\newcounter{paperdefinition}
\newcommand{\paperdefinition}[1]{%
  \refstepcounter{paperdefinition}%
  \par\smallskip\noindent\textbf{\emph{Definition \thepaperdefinition\ (#1):}}\nobreak\hspace{0.3em}}

\definecolor{ink}{HTML}{1F2937}
\definecolor{slate}{HTML}{64748B}
\definecolor{footgray}{HTML}{6B7280}
\definecolor{planfill}{HTML}{E8F0F7}\definecolor{planbord}{HTML}{3D6A96}
\definecolor{neutfill}{HTML}{F1F3F5}\definecolor{neutbord}{HTML}{64748B}
\definecolor{priofill}{HTML}{FBF3DA}\definecolor{priobord}{HTML}{B08A2E}
\definecolor{gatfill}{HTML}{F0EBF7}\definecolor{gatbord}{HTML}{7C5FA8}
\definecolor{safefill}{HTML}{E7F3EA}\definecolor{safebord}{HTML}{2F7D46}
\definecolor{obstfill}{HTML}{FDEBD3}\definecolor{obstbord}{HTML}{C97D1E}
\definecolor{agfill}{HTML}{FBE7E9}\definecolor{agbord}{HTML}{BE3B4A}

\begin{document}
\raggedbottom

\title{CollisionGAT: Controller-Agnostic One-Step Collision Screening for Multi-Agent Motion}

\author{%
\IEEEauthorblockN{Alan Debbas}
\IEEEauthorblockA{\textit{McGill University}\\
Montreal, Canada\\
Alan.Debbas@mail.mcgill.ca}
\and
\IEEEauthorblockN{Edwin Meriaux}
\IEEEauthorblockA{\textit{McGill University}\\
Montreal, Canada\\
Edwin.Meriaux@mail.mcgill.ca}
\and
\IEEEauthorblockN{Gregory Dudek}
\IEEEauthorblockA{\textit{McGill University}\\
Montreal, Canada\\
Gregory.Dudek@mcgill.ca}}

\AddToShipoutPictureBG*{%
  \AtTextLowerLeft{%
    \raisebox{-0.45in}{\parbox[b]{\textwidth}{\scriptsize
    \copyright~2026 IEEE. Personal use of this material is permitted. Permission
    from IEEE must be obtained for all other uses, in any current or future media,
    including reprinting/republishing this material for advertising or promotional
    purposes, creating new collective works, for resale or redistribution to
    servers or lists, or reuse of any copyrighted component of this work in other
    works.}}}}

\maketitle

\begin{abstract}
Before a team of robots moves, each proposed step must be
checked for collisions with other robots and with obstacles.
We present CollisionGAT, a graph-attention network that reads
the current and proposed states of moving agents together with
locally relevant stationary obstacles and returns one
collision-risk score per moving agent. Any controller can use
these scores to accept, repair, replan, or postpone a proposed
step. We mount CollisionGAT on a continuous path-following
controller and on GATeD, an obstacle-blind D$^*$ Lite planner
that uses typed vetoes to update its planning graphs. Exact
geometric checks supply the training labels and independently
audit every executed step.
\end{abstract}

\begin{IEEEkeywords}
multi-agent systems, collision detection, graph attention networks,
motion planning, D* Lite
\end{IEEEkeywords}

\section{Introduction}
Before several robots move together, a controller must decide whether their proposed next steps can be executed without collision.
The decision is relational: a step that is safe for one robot alone can land on another robot's destination, enter an obstacle, or send two robots to the same grid cell.
The number of neighbors that matter changes at each timestamp, so the checker must handle a variable number of agents and obstacles.

Exact geometric predicates supply training labels and audit
execution, but exhaustive screening evaluates up to
$\binom{n}{2}+nm$ agent--agent and agent--obstacle pairs.
Before inference, each proposed motion segment is enclosed in an
axis-aligned box expanded by the agent radius; only sender pairs
whose boxes can overlap are retained. CollisionGAT batches these
pairs without receiving relative-distance or collision-margin
features.

The interface is simple: each agent proposes one next state, CollisionGAT returns one risk score per proposal, and the controller decides whether to commit, repair, replan, or wait.

Our contributions are (i) a complete input--output contract for a four-head pairwise-attention collision classifier, (ii) a training progression from small collision primitives to full task-scale graphs, and (iii) a continuous and a discrete controller built on the same interface with separately trained weights.

\section{Background}

CollisionGAT draws on three lines of work: graph attention
networks, learned safety checking for multi-agent motion, and
incremental path planning.

Graph attention networks learn how much each neighbor should
matter \cite{gat}, and dynamic pairwise attention lets each
receiver rank the same neighbors differently \cite{gatv2}.
CollisionGAT uses this relational machinery for a narrow task:
classifying each agent's proposed step as risky or safe.

Learned models have been used to accelerate collision queries in
motion planning \cite{fastron}. Control-admissibility models
score sampled actions with a graph network while a separate
function drives progress toward the goal \cite{cam}. GCBF+
jointly learns a graph barrier function and a distributed
controller \cite{gcbfplus}. CollisionGAT deliberately does less:
the external controller supplies the action, the network only
classifies it, and masked sender views attribute a grid veto to
an agent or obstacle. Control synthesis and formal certification
remain complementary to this checking role.

Our grid controller builds on D$^*$ Lite, which incrementally
repairs shortest paths when edge costs change \cite{dstarlite}.
Deterministic priority rules resolve simultaneous agent
conflicts, while the frozen checker supplies the online graph
updates.

\section{Problem Formulation}\label{sec:problem-formulation}
\paperdefinition{World and active agents}\label{def:world}
Let $\mathcal W$ be a world containing a finite initial set of mobile agents $\mathcal A_0$ and stationary obstacles $\mathcal O$.
Agent $i$ has start $\bm s_i\in\mathcal W$ and goal $\bm g_i\in\mathcal W$.
At timestamp $t$, the active set $\agents\subseteq\mathcal A_0$ contains the agents that have not yet reached their goals.
Active agent $i$ occupies $\bm p_i$, and the controller proposes a one-step state $\bm q_i$.
Arrival is absorbing: when agent $i$ arrives,
$\mathcal A_t \leftarrow \mathcal A_t\setminus\{i\}$.
Unsubscripted bold symbols such as $\bm p=(\bm p_i)_{i\in\agents}$ stack the indexed states.

In continuous motion, $\mathcal W=[0,L]^2\subset\mathbb R^2$ and $\bm q_i=\bm p_i+\ell_i\bm d_i\in\mathcal W$, where $\ell_i\ge0$ and $\|\bm d_i\|_2=1$ when $\ell_i>0$.
Agent $i$ occupies the closed ball $\overline B(z,r_i):=\{x:\|x-z\|_2\le r_i\}$ with $r_i>0$, and both the current and the proposed footprint lie inside $\mathcal W$.
Each obstacle $o\in\mathcal O$ is a pair $(\bm c_o,a_o)$ with $a_o>0$ and occupies the closed square $Q(\bm c_o,a_o):=\{x:\max(|x_1-c_{o,1}|,|x_2-c_{o,2}|)\le a_o\}$.
In discrete motion, $\mathcal W=\grid:=\{0,\ldots,N-1\}^2$, the obstacle set is a subset of grid cells, starts are pairwise distinct. We define the starting position and end goal of agent $i$ as $\bm s_i,\bm g_i\in\grid\setminus\mathcal O$.
A legal proposal either stays in place or moves to one of the four neighboring cells:
\begin{equation}
\mathcal N_4(c):=\{c'\in\grid:\|c'-c\|_1=1\},\qquad
\bm q_i\in\{\bm p_i\}\cup\mathcal N_4(\bm p_i).
\label{eq:n4}
\end{equation}

\begin{figure}[t]
\centering
\resizebox{0.95\columnwidth}{!}{\begin{tikzpicture}[
  >=Stealth,
  agent/.style={circle,draw=black,fill=blue!25,minimum size=5.5mm,inner sep=0pt,font=\scriptsize},
  prop/.style={->,thick,blue!60!black},
  lbl/.style={font=\scriptsize},
  panelcap/.style={font=\scriptsize\itshape,align=center},
]
% ---------- (a): continuous endpoint overlap ----------
\begin{scope}[shift={(0,0)}]
  \draw[dashed,gray] (0.2,0.4) circle (0.28);
  \draw[dashed,gray] (2.4,1.6) circle (0.34);
  \fill[blue!20,draw=blue!60!black] (1.25,1.05) circle (0.28);
  \fill[red!20,draw=red!60!black]   (1.65,1.15) circle (0.34);
  \draw[prop] (0.2,0.4) -- (1.05,0.92);
  \draw[prop,red!60!black] (2.4,1.6) -- (1.9,1.28);
  \node[lbl] at (0.2,0.02) {$\bm p_i$};
  \node[lbl] at (2.4,2.02) {$\bm p_j$};
  \node[lbl] at (0.95,1.42) {$\bm q_i$};
  \node[lbl] at (2.15,0.85) {$\bm q_j$};
  \node[panelcap] at (1.3,-0.62) {(a) endpoint overlap:\\detected by swept label};
\end{scope}
% ---------- (b): continuous crossing, safe endpoints ----------
\begin{scope}[shift={(4.1,0)}]
  \draw[dashed,gray] (0.2,1.6) circle (0.28);
  \draw[dashed,gray] (0.2,0.3) circle (0.28);
  \fill[blue!20,draw=blue!60!black] (2.3,0.3) circle (0.28);
  \fill[red!20,draw=red!60!black]   (2.3,1.6) circle (0.28);
  \draw[prop] (0.2,1.6) -- (2.0,0.42);
  \draw[prop,red!60!black] (0.2,0.3) -- (2.0,1.48);
  \node[font=\footnotesize,red!70!black] at (1.25,0.95) {$\times$};
  \node[lbl] at (0.2,2.0) {$\bm p_i$};
  \node[lbl] at (0.2,-0.12) {$\bm p_j$};
  \node[lbl] at (2.3,-0.12) {$\bm q_i$};
  \node[lbl] at (2.3,2.0) {$\bm q_j$};
  \node[panelcap] at (1.25,-0.62) {(b) crossing, safe endpoints:\\detected by swept label};
\end{scope}
% ---------- (c): grid duplicate destination ----------
\begin{scope}[shift={(0.3,-4.0)}]
  \draw[gray!60] (0,0) grid (2,2);
  \node[agent] (A) at (0.5,1.5) {$i$};
  \node[agent,fill=red!25] (B) at (1.5,0.5) {$j$};
  \draw[prop] (A) -- (1.32,1.42);
  \draw[prop,red!60!black] (B) -- (1.5,1.28);
  \draw[red!70!black,thick] (1.06,1.06) rectangle (1.94,1.94);
  \node[lbl] at (1.0,2.28) {$\bm q_i=\bm q_j$};
  \node[panelcap] at (1.0,-0.62) {(c) duplicate destination:\\detected by grid label};
\end{scope}
% ---------- (d): grid head-on swap ----------
\begin{scope}[shift={(4.6,-4.0)}]
  \draw[gray!60] (0,0) grid (2,2);
  \node[agent] (C) at (0.5,1.0) {$i$};
  \node[agent,fill=red!25] (D) at (1.5,1.0) {$j$};
  \draw[prop] (C.north) .. controls (0.7,1.55) and (1.3,1.55) .. (D.north);
  \draw[prop,red!60!black] (D.south) .. controls (1.3,0.45) and (0.7,0.45) .. (C.south);
  \node[font=\footnotesize,red!70!black] at (1.0,1.0) {$\times$};
  \node[lbl] at (1.0,2.28) {$\bm q_i=\bm p_j,\ \bm q_j=\bm p_i$};
  \node[panelcap] at (1.0,-0.62) {(d) head-on swap:\\detected by grid label};
\end{scope}
\end{tikzpicture}}
\caption{Agent--agent collision cases in continuous and grid
motion. Dashed circles indicate current positions and solid
discs or cells indicate proposed positions. The continuous
swept label detects (a) endpoint overlap (subcase of a crossing overlap) and (b) crossing
between safe endpoints. Under four-connected grid motion,
(c) duplicate destinations and (d) head-on swaps are the only
possible one-step agent collisions; every participant is labeled.}
\label{fig:labels}
\end{figure}
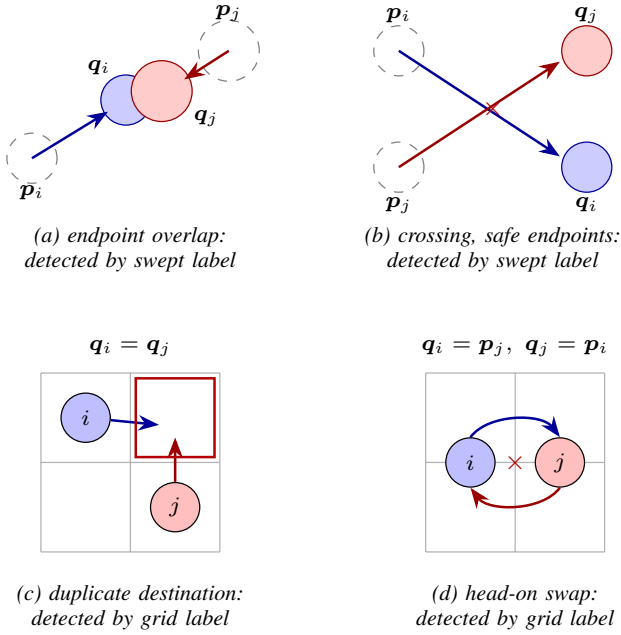
The next two definitions give the exact collision checks.
They supply all training labels and independently re-check every executed step; Fig.~\ref{fig:labels} illustrates the agent cases.
Throughout, $\ind\{P\}=1$ when proposition $P$ is true and $0$ otherwise, and a maximum over an empty set is zero.

\paperdefinition{Grid labels}\label{def:grid-collision}
A step collides when it enters an obstacle cell, or when two agents claim the same cell or swap head-on (Fig.~\ref{fig:labels}c,d):
\begin{align}
y_i^{\rm obs}&=\ind\{\bm q_i\in\mathcal O\},\label{eq:grid-obstacle}\\
y_i^{\rm agt}&=\ind\!\left\{\exists j\in\agents\setminus\{i\}:\right.\notag\\[-1mm]
&\hspace{8mm}\left.\bm q_i=\bm q_j\ \vee\
(\bm q_i=\bm p_j\wedge\bm q_j=\bm p_i)\right\}.
\label{eq:grid-agent}
\end{align}
For the grid model, $y_i^{\rm all}=y_i^{\rm agt}\vee y_i^{\rm obs}$.

\paperdefinition{Swept-motion labels}\label{def:swept}
In continuous motion, agent \(i\) follows the straight center
trajectory
\[
\gamma_i(\lambda)
=(1-\lambda)\bm p_i+\lambda\bm q_i,
\qquad \lambda\in[0,1].
\]
Its swept agent-collision label is
\begin{equation}
y_i^{\mathrm{sw,agt}}
=
\max_{j\in\agents\setminus\{i\}}
\ind\!\left\{
\min_{\lambda\in[0,1]}
\|\gamma_i(\lambda)-\gamma_j(\lambda)\|_2
\le r_i+r_j
\right\}.
\label{eq:swept-agent}
\end{equation}
For a point \(\bm x\) and a closed square \(Q\), let
\(\operatorname{dist}(\bm x,Q)
=\min_{\bm z\in Q}\|\bm x-\bm z\|_2\).
The swept obstacle-collision label is
\begin{equation}
y_i^{\mathrm{sw,obs}}
=
\max_{o\in\mathcal O}
\ind\!\left\{
\min_{\lambda\in[0,1]}
\operatorname{dist}
\bigl(\gamma_i(\lambda),Q(\bm c_o,a_o)\bigr)
\le r_i
\right\}.
\label{eq:swept-obstacle}
\end{equation}
For the continuous model,\[y_i^{\mathrm{sw,all}}=y_i^{\mathrm{sw,agt}}\lor
y_i^{\mathrm{sw,obs}}.\]

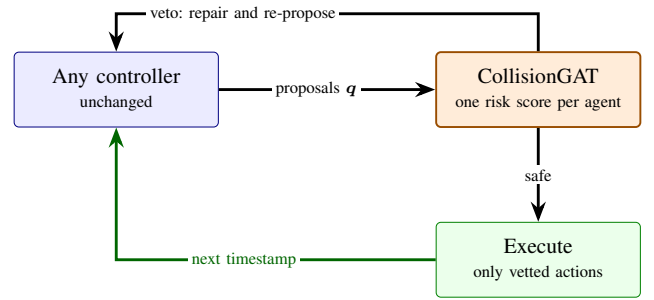
\begin{figure}[b]
\centering
\resizebox{0.95\columnwidth}{!}{\begin{tikzpicture}[
  >=Stealth,
  font=\small,
  box/.style={draw,rounded corners=2pt,align=center,inner sep=5pt,minimum height=11mm},
  ctrl/.style={box,fill=blue!8,draw=blue!50!black,minimum width=30mm},
  exec/.style={box,fill=green!8,draw=green!50!black,minimum width=30mm},
  gat/.style={box,fill=orange!15,draw=orange!60!black,thick,minimum width=30mm},
  flow/.style={->,very thick},
  edgelbl/.style={font=\scriptsize,align=center,fill=white,inner sep=1.5pt},
]
\node[ctrl] (c) at (0,0) {Any controller\\\scriptsize unchanged};
\node[gat]  (g) at (6.2,0) {CollisionGAT\\\scriptsize one risk score per agent};
\node[exec] (e) at (6.2,-2.5) {Execute\\\scriptsize only vetted actions};
\draw[flow] (c.east) -- node[edgelbl,pos=0.45]{proposals $\bm q$} (g.west);
\draw[flow] (g.south) -- node[edgelbl]{safe} (e.north);
\draw[flow] (g.north) -- ++(0,0.55) -| node[edgelbl,pos=0.35]{veto: repair and re-propose} (c.north);
\draw[flow,green!40!black] (e.west) -| node[edgelbl,pos=0.3]{next timestamp} (c.south);
\end{tikzpicture}}
\caption{CollisionGAT attaches between any controller and execution: safe
joint proposals pass through, and a veto sends the controller back to
repair and re-propose. The controller itself is unchanged.}
\label{fig:attach}
\end{figure}
\section{CollisionGAT}
CollisionGAT sits in conjunction with an external controller.
The controller sends in the current and proposed states of every node, each typed as an agent or an obstacle, and CollisionGAT returns one collision-risk score per active agent.
This section answers three questions: what the network looks at (the interaction graph), what numbers describe each entity (the features), and how those numbers become a risk score (the classifier).
This model can be seen in Fig.~\ref{fig:attach}.
 
\subsection{Interaction graph and minimal features}
 
\paperdefinition{Continuous and grid interaction graphs}
\label{def:graph}
Let $\chi_t(u,i)$ be a binary edge filter. We set
$\chi_t(u,i)=1$ when sender $u$ could collide with agent $i$
during the proposed step, and $\chi_t(u,i)=0$ otherwise. Let $\ell_{\max}$ be the largest continuous candidate step. For continuous motion,
\begin{align}
\mathcal S_t^{\rm cont}
&=
\left\{
o\in\mathcal O:
\substack{
\exists i\in\agents,\\
\operatorname{dist}(\bm p_i,Q(\bm c_o,a_o))
\leq\ell_{\max}+r_i
}
\right\}, \nonumber\\
\mathcal V_t^{\rm cont}
&=
\agents\sqcup\mathcal S_t^{\rm cont},
\nonumber\\
\mathcal E_t^{\rm cont}
&=
\left\{
(u,i):
\substack{
i\in\agents,\\
u\in\mathcal V_t^{\rm cont}\setminus\{i\},\
\chi_t(u,i)=1
}
\right\}.
\label{eq:continuous-graph}
\end{align}

For grid motion,
\begin{align}
\mathcal S_t^{\rm grid}
&=
\mathcal O\cap
\bigcup_{i\in\agents}
\left(\{\bm p_i\}\cup\mathcal N_4(\bm p_i)\right),
\nonumber\\
\mathcal V_t^{\rm grid}
&=
\agents\sqcup\mathcal S_t^{\rm grid},
\nonumber\\
\mathcal E_t^{\rm grid}
&=
\left\{
(u,i):
\substack{
i\in\agents,\\
u\in\mathcal V_t^{\rm grid}\setminus\{i\}
}
\right\}.
\label{eq:grid-graph}
\end{align}

\paperdefinition{Raw states and sender masks}
For continuous scenes, let $\bm c_t$ be the centroid (average
position) of the active-agent positions. The raw
seven-dimensional agent and obstacle rows are
\begin{equation}
\begin{aligned}
\bm x_i
&=[\bm p_i-\bm c_t\Vert\bm q_i-\bm c_t
  \Vert\ell_i\Vert r_i\Vert0],\\[-0.5mm]
\bm x_o
&=[\bm c_o-\bm c_t\Vert\bm c_o-\bm c_t
  \Vert0\Vert a_o\Vert1].
\end{aligned}
\label{eq:raw7}
\end{equation}
Each row contains the centered current and proposed positions,
step length, entity size, and a binary obstacle indicator.
Obstacles are stationary, so their positions are repeated and
their step length is zero. Centering removes global translation.

For a grid, let
$\bm c_N=((N-1)/2,(N-1)/2)$ and
$e_N(c)=(c-\bm c_N)/50$. The origin may lie between cells
when $N$ is even; the divisor is half the largest trained side
and only stabilizes magnitude. The raw five-dimensional rows are
\begin{equation}
\bm x_i=[e_N(\bm p_i)\Vert e_N(\bm q_i)\Vert0],
\qquad
\bm x_o=[e_N(o)\Vert e_N(o)\Vert1].
\label{eq:raw5}
\end{equation}
Neither model receives engineered relative displacement,
distance, or collision-margin features. Stacking the rows gives
$X_t^{\rm cont}\in
\mathbb R^{|\mathcal V_t^{\rm cont}|\times7}$ or
$X_t^{\rm grid}\in
\mathbb R^{|\mathcal V_t^{\rm grid}|\times5}$; we write $X_t$
when the setting is clear.
 
\subsection{Four-head pairwise-attention classifier}
\paperdefinition{CollisionGAT map}We follow standard graph-attention networks terminology and theory \cite{gat,gatv2}. For a receiving agent $i$ and
a sending node $u$, the pair representation is
$\bm z_{ui}=[\bm x_u\Vert\bm x_i]\in\mathbb R^{2d}$, where
$d\in\{5,7\}$. For each attention head
$h\in\{1,\ldots,4\}$, the learned MLP
$\phi_a^h:\mathbb R^{2d}\rightarrow\mathbb R$ maps
$\bm z_{ui}$ to a scalar attention logit, while
$\phi_m^h:\mathbb R^{2d}\rightarrow\mathbb R^{16}$ maps it
to the message sent from $u$ to $i$.
For a sender view
$\mu\in\{\mathrm{all},\mathrm{agt},\mathrm{obs}\}$, define:
\[
\begin{aligned}
\mathcal E_t^{\rm all}&=\mathcal E_t,\\
\mathcal E_t^{\rm agt}
&=\{(u,i)\in\mathcal E_t:u\in\agents\},\\
\mathcal E_t^{\rm obs}
&=\{(u,i)\in\mathcal E_t:u\in\mathcal S_t\},
\end{aligned}
\qquad
\mathcal N_i^\mu=\{u:(u,i)\in\mathcal E_t^\mu\}.
\]
If the set is empty, set $\bm M_i^\mu=\bm0$; otherwise, each head takes the attention-weighted average of its messages:
\begin{equation}
\bm M_i^\mu=\mathop{\Vert}_{h=1}^{4}\sum_{u\in\mathcal N_i^\mu}
\frac{e^{\phi_a^h(\bm z_{ui})}}
{\sum_{v\in\mathcal N_i^\mu}e^{\phi_a^h(\bm z_{vi})}}
\phi_m^h(\bm z_{ui})\in\mathbb R^{64}.
\label{eq:attention}
\end{equation}
Different heads may emphasize different interactions. Let
$\phi_x:\mathbb R^d\to\mathbb R^{64}$ be the self encoder,
$\phi_u:\mathbb R^{128}\to\mathbb R^{64}$ the update map, and
$\bm w\in\mathbb R^{64}$, $b\in\mathbb R$ the shared
classifier parameters. Using layer normalization
$\operatorname{LN}$ and the logistic sigmoid
$\sigma(z)=(1+e^{-z})^{-1}$, the output is:
\begin{equation}
\begin{aligned}
\bm h_i^\mu&=\operatorname{LN}\!\left(
\phi_u([\phi_x(\bm x_i)\Vert\bm M_i^\mu])+\phi_x(\bm x_i)\right),\\
\hat y_i^\mu&=\sigma(\bm w^\top\bm h_i^\mu+b).
\end{aligned}
\label{eq:classifier}
\end{equation}
Let $\theta$ collect the trainable parameters.
Then
\begin{equation}
\operatorname{GAT}_\theta(X_t,\mathcal E_t^\mu)
=(\hat y_i^\mu)_{i\in\agents}
\label{eq:gat-map}
\end{equation}
maps one graph view to one risk score per active agent; obstacle nodes are never scored.
 
The self, message, and update maps are two-layer ReLU MLPs, and each attention map has one ReLU hidden layer and a single scalar output.
In both instances, hidden layers have width $64$, each head
outputs $16$ message features, and dropout is $0.10$.
The continuous model has $29{,}381$ trainable parameters; the grid model has $27{,}205$.
 
\paperdefinition{Collision warnings}
Let $\hat c_i\in\{0,1\}$ be the checker decision for agent $i$:
$\hat c_i=1$ warns against its proposal, while $\hat c_i=0$
allows it to pass. The deployed decisions are:
\[
\begin{aligned}
\hat c_i^{\rm cont}
&=\ind\{\hat y_i^{\rm agt}\geq\tau_{\rm agt}
\lor\hat y_i^{\rm obs}\geq\tau_{\rm obs}\},\\
\hat c_i^{\rm grid}
&=\ind\{\hat y_i^{\rm all}\geq\tau_{\rm all}
\lor\hat y_i^{\rm obs}\geq\tau_{\rm obs}\}.
\end{aligned}
\]
All views share one weight set, and their scores are tuned risks,
not calibrated probabilities. The grid agent-only view is used
only for diagnosis.
 
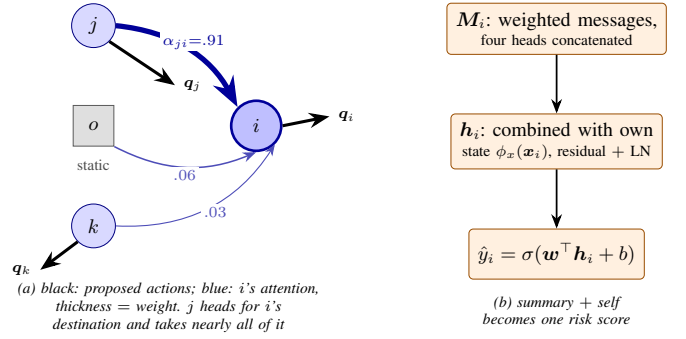
\begin{figure}[t]
\centering
\resizebox{\columnwidth}{!}{\begin{tikzpicture}[
  >=Stealth,
  font=\small,
  agentn/.style={circle,draw=blue!50!black,fill=blue!15,minimum size=7mm,inner sep=0pt},
  recv/.style={circle,draw=blue!60!black,very thick,fill=blue!25,minimum size=8mm,inner sep=0pt},
  obsn/.style={draw=gray!70!black,fill=gray!25,minimum size=6.5mm,inner sep=0pt},
  stage/.style={draw,rounded corners=2pt,align=center,fill=orange!12,draw=orange!60!black,inner sep=4pt,minimum height=8mm},
  act/.style={->,line width=1.4pt,black},
  lbl/.style={font=\scriptsize},
  wlbl/.style={font=\scriptsize,fill=white,inner sep=1pt},
  panelcap/.style={font=\scriptsize\itshape,align=center},
]
% ===== left panel: actions first, then i's attention over its senders =====
\begin{scope}
  \node[recv]  (i) at (2.6,0)    {$i$};
  \node[agentn](j) at (0,1.6)    {$j$};
  \node[agentn](k) at (0,-1.6)   {$k$};
  \node[obsn]  (o) at (0,0)      {$o$};
  % --- proposed actions (black): j heads for i's destination, k departs, o static, i steps right
  \draw[act] (i.east) -- ++(0.75,0.15) node[lbl,right]{$\bm q_i$};
  \draw[act] (j.south east) -- ++(1.05,-0.70) node[lbl,right]{$\bm q_j$};
  \draw[act] (k.south west) -- ++(-0.60,-0.45) node[lbl,left]{$\bm q_k$};
  \node[lbl,gray!60!black] at (0,-0.62) {static};
  % --- attention (blue arcs into i): thickness = weight
  \draw[->,line width=2.4pt,blue!60!black] (j.east) to[bend left=28]
    node[wlbl,pos=0.55,above]{$\alpha_{ji}{=}.91$} (i.north west);
  \draw[->,line width=0.7pt,blue!60!black!70] (o.south east) to[bend right=25]
    node[wlbl,pos=0.5,below]{$.06$} (i.south);
  \draw[->,line width=0.5pt,blue!60!black!70] (k.east) to[bend right=30]
    node[wlbl,pos=0.55,below]{$.03$} (i.south east);
  \node[panelcap] at (1.2,-2.9) {(a) black: proposed actions; blue: $i$'s attention,\\thickness $=$ weight. $j$ heads for $i$'s\\destination and takes nearly all of it};
\end{scope}
% ===== right panel: score pipeline =====
\begin{scope}[shift={(7.4,1.55)}]
  \node[stage] (m)  at (0,0)    {$\bm M_i$: weighted messages,\\[-1pt]\scriptsize four heads concatenated};
  \node[stage] (u)  at (0,-1.8) {$\bm h_i$: combined with own\\[-1pt]\scriptsize state $\phi_x(\bm x_i)$, residual $+$ LN};
  \node[stage] (y)  at (0,-3.6) {$\hat y_i=\sigma(\bm w^\top\bm h_i+b)$};
  \draw[->,thick] (m) -- (u);
  \draw[->,thick] (u) -- (y);
  \node[panelcap] at (0,-4.5) {(b) summary $+$ self\\becomes one risk score};
\end{scope}
\end{tikzpicture}}
\caption{CollisionGAT in one picture. (a) Black arrows are proposed actions;
blue arcs are receiver $i$'s attention over its senders
(Eq.~\ref{eq:attention}), with thickness proportional to weight. Agent $j$ proposes $i$'s destination and is emphasized by the
illustrative attention weights. (b) The four-head summary is combined with the
agent's own state and mapped to one risk score (Eq.~\ref{eq:classifier}).}
\label{fig:arch}
\end{figure}
 
\subsection{Training and collision-detection results}

\emph{Discrete model.}
The grid model first learned elementary collision patterns from
40,000 primitive scenes and was then fine-tuned on 2,100
large-world joint-transition scenes. Its held-out test set contained 1,680 joint proposals and
8,364 agent decisions. It achieved precision, recall, and F1 of
.9975, 1.0000, and .9987.

\emph{Continuous model.}
The swept-motion model began with 2,800 primitive graphs.
Task adaptation used 4,410 goal-directed graphs and 1,102
primitive replay graphs. A further 13,200 deployment-matched, difficult, and controller-generated graphs refined the frozen model. The frozen test contained 1,000 joint proposals
from 100 new seed-disjoint world groups, producing 9,950
agent decisions. It achieved precision, recall, and F1 of
.9127, .9996, and .9542.
  
\section{Algorithm 1: Continuous Motion}

Algorithm~1 mounts the frozen CollisionGAT on a continuous
waypoint controller. Before each episode, every agent
independently runs deterministic A$^*$ on a four-connected
lattice constructed from the known static-obstacle map
\cite{astar}. The resulting path is greedily shortened into an
ordered waypoint sequence
$\mathcal W_i=(\bm w_{i,1},\ldots,\bm w_{i,K_i}=\bm g_i)$,
with every retained straight segment checked for
disc--obstacle clearance. During execution, the controller
targets the current waypoint and advances to the next one when
it is reached.

At timestamp $t$, agent $i$ samples
$\ell_{i,t}\sim U[.4,1.2]$ and proposes a step toward its current
waypoint $\bm w_{i,t}$:
\begin{equation}
\bm q_i=\bm p_i+
\min\!\left\{\ell_{i,t},
\|\bm w_{i,t}-\bm p_i\|_2\right\}
\frac{\bm w_{i,t}-\bm p_i}
{\|\bm w_{i,t}-\bm p_i\|_2}.
\label{eq:direct}
\end{equation}

The minimum shortens the final step so that the agent does not
pass its waypoint. A moving proposal is \emph{unsafe} if CollisionGAT warns it directly or if it can reach a warned waiting receiver in one step. \textbf{Fair priority} favors the longest-waiting agent. If their waiting times are equal, index priority rotates deterministically across timestamps. Algorithm~1 commits no GAT-positive moving proposal.

We can showcase CollisionGAT because the obstacle-clear routes are planned independently,
they do not coordinate interactions between agents. 
\begin{algorithm}[t]
\caption{CollisionGAT-Guard (Continuous)}
\label{alg:continuous}
\begin{algorithmic}[1]
\While{$\mathcal A_t\neq\emptyset$ and the episode limit is unmet}
  \State propose $q_i$ using \eqref{eq:direct} for every active $i$;
  resolve shared targets by \textbf{fair priority}
  \State assess $\bm q$ with the frozen $\mathrm{GAT}_\theta$
  \While{an agent is warned and fewer than five repairs were tried}
    \State reduce one warned agent's speed; reassess $\bm q$
  \EndWhile
  \State hold remaining warned agents; restore one at a time
   only when reassessment gives no moving warning
  \If{a moving warning remains} \State $\bm q\gets\bm p$ \EndIf
  \State commit $\bm p\gets\bm q$ synchronously; retire arrivals
\EndWhile
\end{algorithmic}
\end{algorithm}

\begin{figure*}[t]
\centering
\includegraphics[
  width=0.8\textwidth
]{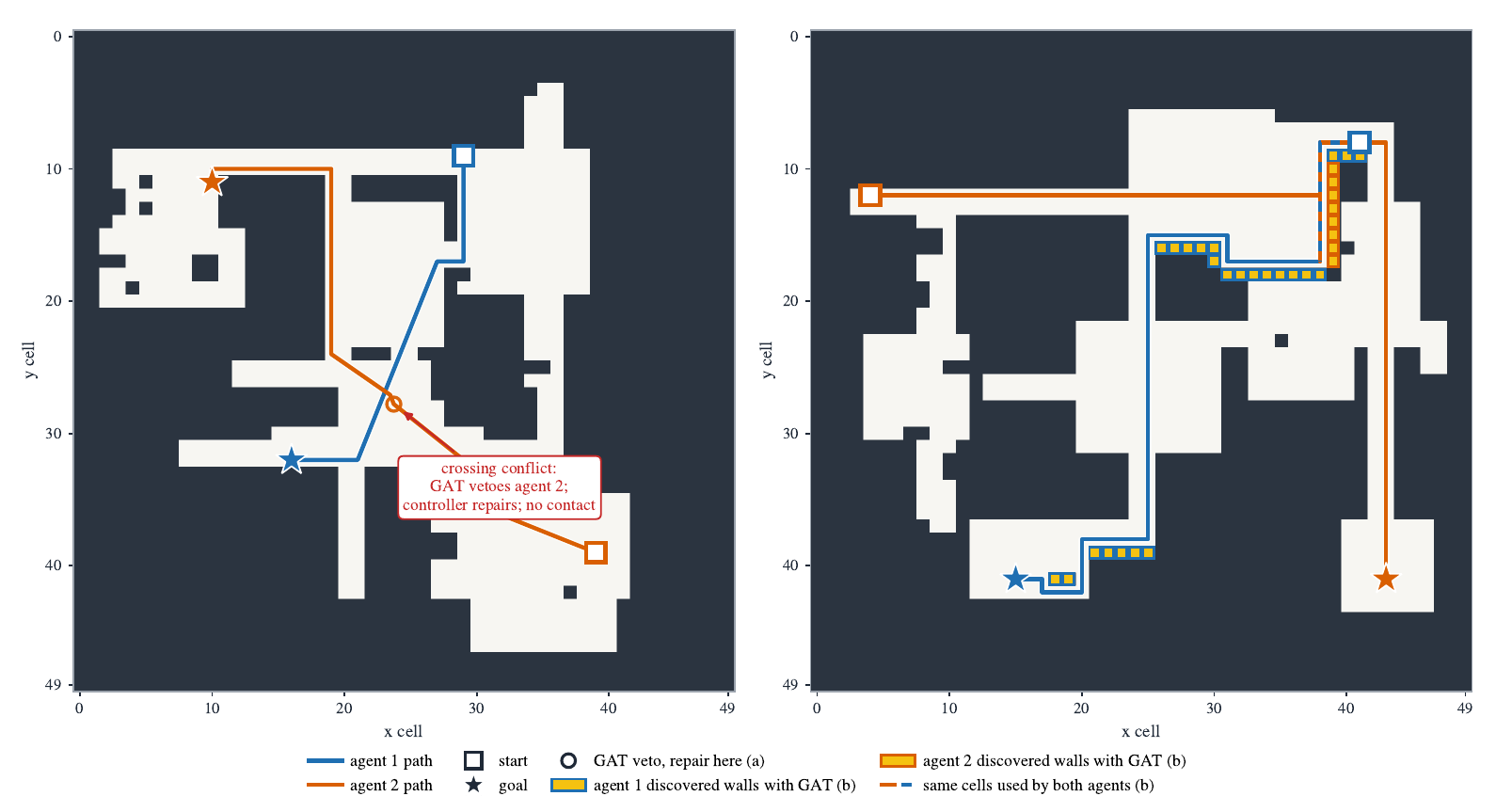}
\caption{Full trajectories of two CollisionGAT mounts.
Left: independently planned continuous waypoint routes create
an agent--agent crossing conflict; CollisionGAT vetoes the
proposal before controller repair. Right: obstacle-blind GATeD
uses obstacle-attributed vetoes to discover and share wall cells
before repairing D$^*$ Lite paths. Squares and stars denote
starts and goals.}
\label{fig:mounts}
\end{figure*}

\section{Algorithm 2: GATeD}
The discrete Algorithm~2, \emph{GATeD}, separates planning from
checking. Agent $i$ owns a complete four-connected grid graph
$G_{i,t}$ with unit-cost available edges. The planner never sees
the true obstacle set; CollisionGAT validates each joint proposal
using only the local obstacle senders of
Definition~\ref{def:graph} and returns typed warnings.

\paragraph{Planning}
$\mathrm D^*(G_{i,t},p_i,g_i)$ incrementally
repairs agent $i$'s D$^*$ Lite state and returns its next
minimum-cost cell, or $p_i$ at the goal or when no path exists.
The shared discovered-wall set $\widehat{\mathcal O}$ starts
empty. Private edge restrictions expire at the next timestamp,
whereas shared wall restrictions persist.

\paragraph{Vetoes}
The known-free set $\mathcal K$ initially contains every start
and goal and gains every cell occupied after a committed move.
An obstacle-attributed warning on $q_i\notin\mathcal K$ adds
$q_i$ to $\widehat{\mathcal O}$ and permanently removes that
cell's incident edges from every planner. Every other moving GAT
warning or priority loss removes only $(p_i,q_i)$ from agent
$i$'s graph until the next timestamp. If the same active agents
return to the same positions without a new wall discovery, the
controller cyclically selects one moving agent, temporarily
blocks its next D$^*$ Lite edge, and validates the replanned
proposal normally.

\paragraph{Decisions}
Before learned warnings are acted upon, deterministic priority
resolves duplicate destinations, swaps, and moves into cells held
by waiting agents: a stationary occupant wins, otherwise the
lowest-ID agent wins. Agents replan until a joint proposal is
accepted or no repair remains; otherwise all agents wait for that
timestamp. Exact grid predicates audit committed steps but never
modify an action.  
 
\begin{algorithm}[t]
\caption{GATeD (Discrete)}
\label{alg:gated}
\begin{algorithmic}[1]
\While{$\mathcal A_t\neq\emptyset$}
  \State restore private edges; $\mathit{accepted}\gets\mathrm{False}$
  \While{not accepted and the repair budget remains}
    \State $q_i\gets\mathrm D^*(G_{i,t},p_i,g_i)$ for every active $i$;
    assess $\bm q$ with the frozen $\mathrm{GAT}_\theta$
    \If{duplicate destinations or swaps exist}
      \State temporarily remove each losing agent’s edge \((p_i,q_i)\) from its own graph until the next timestamp
    \ElsIf{a moving proposal is warned}
      \State permanently remove an obstacle-vetoed cell from every agent’s graph; otherwise remove only the warned agent’s edge \((p_i,q_i)\) until the next timestamp
    \Else
      \State commit $\bm p\gets\bm q$; update $\mathcal K$;
      retire arrivals; $\mathit{accepted}\gets\mathrm{True}$
    \EndIf
  \EndWhile
  \If{not accepted} \State $\bm q\gets\bm p$ 
  \EndIf
\EndWhile
\end{algorithmic}
\end{algorithm}

The exact grid checks independently audit every committed step.
Each planner keeps D$^*$ Lite's incremental updates, while deterministic priority and temporary repairs coordinate the joint action.

\section{Evaluation Results}

We evaluate the continuous and discrete mounts on separate
100-world test suites. For the PNG maps, structural complexity
is measured by the number of nodes in an occupancy quadtree,
where fragmented free--obstacle boundaries require further
subdivision \cite{meriaux2025}. The evaluated maps span three
consecutive complexity bands: $15{,}000$--$26{,}999$,
$27{,}000$--$38{,}999$, and $39{,}000$--$50{,}999$
quadtree nodes.

The continuous suite combines 50 generated $25\times25$
worlds containing square obstacles with 50 $50\times50$ PNG
maps from the lowest complexity band. Each episode contains
7--13 agents assigned to one to three goals. Of 994 agents,
846 arrived, and 66 of the 100 episodes completed. The
remaining 34 episodes stalled with one or more agents short of
their goals, showing that the waypoint controller does not
guarantee completion. The left panel of
Fig.~\ref{fig:mounts} illustrates CollisionGAT vetoing an
agent--agent crossing before controller repair. No GAT-positive
moving proposal was committed, and the exact swept-motion audit
recorded no executed collision.

\begin{table}[t]
\centering
\caption{Controller outcomes on the two 100-world test suites.}
\label{tab:controller}
\setlength{\tabcolsep}{4pt}
\begin{tabular}{lrrr}
\toprule
Setting & \# Worlds & Agents arrived & Collisions\\
\midrule
Discrete GATeD & 100 & 995/995 & 0\\
Continuous CollisionGAT-Guard & 100 & 846/994 & 0\\
\bottomrule
\end{tabular}
\end{table}

The discrete suite contains 34 maps from the lowest complexity
band, 33 from the middle band, and 33 from the highest band.
It includes 49 $50\times50$ and 51 $100\times100$ maps,
with 7--13 agents per episode and an even split between
shared- and multiple-goal settings. Every episode completed,
all 995 agents arrived, and no executed collision was recorded.
Across 450{,}523 active-agent assessments, the obstacle view
detected all 34{,}703 proposed wall entries with only one false
positive. These vetoes accumulated 27{,}477 true obstacle-cell
discoveries in the agents' shared planning memory.

\section{Discussion and Conclusion}
The main finding is that CollisionGAT can be mounted on substantially different controllers through the same propose--validate--update interface. The continuous waypoint controller and the obstacle-blind D$^*$ Lite planner use different state spaces, planning mechanisms, and responses to rejected actions, yet both provide CollisionGAT only with typed current and proposed states, and both recorded no audited executed collision across their respective 100-world evaluations. These results support portability at the interface level: a proposal-based controller that can repair, replan, or postpone a rejected action can use the checker without exposing its internal planning state. Because the two mounts use separately trained weights, this claim concerns the shared interface and architecture, not transfer of one checkpoint between motion domains.

The arrival results distinguish collision screening from navigation completeness. The continuous controller delivered 846/994 agents, whereas GATeD delivered 995/995 (Table~\ref{tab:controller}); these are not accuracy comparisons because the controllers and test suites differ. The continuous mount may stall after conflicts or conservative warnings, while GATeD updates its planning graphs and retains discovered obstacle information. CollisionGAT screens proposed transitions; the controller remains responsible for progress, deadlock resolution, and arrival.

Typed vetoes make the interface actionable: in GATeD, an obstacle-attributed veto adds the proposed cell to permanent shared wall memory, whereas an agent or priority conflict creates a private one-timestamp restriction, so CollisionGAT guides the controller's response without receiving its planner state. Exact geometric predicates generate the training labels and independently audit every committed transition, so the results do not claim that CollisionGAT is more accurate than exact geometry or that it provides a formal safety certificate. Its value is its reusable relational interface: one frozen checker per motion domain converts a variable-size set of local agent and obstacle interactions into per-agent risk scores without receiving goals, routes, maps, or planner states. Geometry acts as teacher and evaluator, while CollisionGAT provides the learned validation layer mounted between proposal and execution.

This collision-screening result is one instance of a broader idea: the same controller-agnostic validation interface can be mounted on other algorithms and extended to graph attention networks that veto actions violating two other important multi-agent requirements, connectivity and coverage.

\section*{Acknowledgment}
OpenAI Codex and Anthropic Claude Code assisted with software
refactoring, figure preparation, and language editing; the authors
reviewed and verified all outputs.

\end{document}